\documentclass[10pt,twocolumn,letterpaper]{article}

\usepackage{wacv}
\usepackage{times}
\usepackage{epsfig}
\usepackage{graphicx}
\usepackage{amsmath}
\usepackage{amssymb}
\usepackage{multirow}
\usepackage{caption} 
\usepackage{subcaption}

\graphicspath{{./figures/}}
\usepackage[pagebackref=true,breaklinks=true,letterpaper=true,colorlinks,bookmarks=false]{hyperref}

\wacvfinalcopy % *** Uncomment this line for the final submission

\ifwacvfinal\fi
\begin{document}

%%%%%%%%% TITLE
\title{Inverse Visual Question Answering with Multi-Level Attentions }

\author{
	{\bf Yaser Alwattar}\quad and \quad {\bf Yuhong Guo} \\	
School of Computer Science, 
Carleton University, Canada \\
{\tt\small yaseralwattar@cmail.carleton.ca, yuhong.guo@carleton.cca}
%\and
}

\maketitle
\ifwacvfinal\thispagestyle{empty}\fi

%%%%%%%%% ABSTRACT
\begin{abstract}
Inverse Visual Question Answering (iVQA) is a contemporary task emerged from the need of improving visual and language understanding. 
It tackles the challenging problem of generating a corresponding question for a given image-answer pair. 
In this paper, we propose a novel deep multi-level attention model to address 
inverse visual question answering. 
The proposed model generates regional visual and semantic features at the object level
and then enhances them with the answer cue by using attention mechanisms. 
Two levels of multiple attentions are employed in the model, 
including the dual attention at the partial question encoding step 
and the dynamic attention at the question's next word generation step.
We evaluate the proposed model on the VQA V1 dataset. 
It demonstrates 
state-of-the-art performance 
in terms of multiple commonly used metrics. 
\end{abstract}     

%%%%%%%%% BODY TEXT
\section{Introduction}
Inverse Visual Question Answering (iVQA) \cite{r41,r29} is a new interesting problem 
emerged in computer vision 
after the great success of deep learning on the VQA task \cite{r40, r44, r57, r59}. 
iVQA can be viewed as an extension to 
the standard task of Visual Question Generation (VQG) \cite{r6, r24, r23}.
However, different from the VQG task,
which automatically generates general questions from given images,
iVQA restricts the question generation to be correspondent to a pair of image and answer. 
It has been noticed that a VQA system could predict correct answers merely based
on the textual questions without considering the corresponding images \cite{r63},
which indicates the limitation of VQA on image understanding. 
By contrast, an answer phrase is typically much shorter than the question,
it is necessary to integrate both visual image understanding and language semantic understanding
to yield an effective iVQA system.
This makes iVQA more challenging than VQA as both image and answer, 
as well as their complex interactions, 
must be captured. 
It provides a platform for enhancing machine visual and textual understandings.
Meanwhile iVQA systems 
can be used as inexpensive tools to generate questions and create richer datasets for VQA. 

As a newly emerged task, iVQA has been studied in a few previous works~\cite{r29,r41} since its introduction in~\cite{r29}.
The work in~\cite{r29} have tackled iVQA
by enhancing the answer cue with high-level semantic information
only at the initial step of question generation.
The question generation model then dynamically attends to important parts of a given image 
based on the current generation state and the answer cue.
The follow-up work in~\cite{r41} further uses 
variational autoencoders to enhance question diversity of iVQA.
In these works,
the complex interaction and consistent alignment of the textual semantic features with
useful object level visual features, which are important for
effective attentions, however remain unexplored. 
The potential selection of regions that are not informative or aligned with objects
may negatively affect the question generation.

In this paper, we propose a novel multi-level attention model to address inverse visual question answering.
The proposed model induces consistent regional textual and visual feature alignment with the answer cues 
via attention mechanisms. 
It extends a bottom-up attention model \cite{r44} to extract visual features 
at the level of objects and salient regions, as well as generate object and attribute labels, 
i.e., semantic features (or textual features). 
Different from previous iVQA models \cite{r29, r41} that 
simply integrate the answer and high-level semantic concepts
to provide an initial glimpse for the partial question encoder, 
we deploy a dual guiding attention module to attend to 
the relevant object-level parts of both visual features and semantic features
based on the answer cue for partial question encoding.
At the question generation level,
we first statically attend the concatenation of visual and semantic features 
with the answer cue, and then deploy 
a dynamic attention module to attend to important regions 
for the next question word generation. 
We expect such a multi-level attention based model is able to 
generate visual informative and answer-consistent questions 
by enforcing a correct alignment between the visual features
and textual answer cues and achieve better image and text understanding.
To validate the effectiveness of the proposed model,
we conduct experiments on the VQA V1 dataset \cite{r17}.
The results show that our proposed model achieves the state-of-the-art performance 
in terms of many benchmark metrics comparing with other competitors \cite{r29, r41}.

%-------------------------------------------------------------------------}
\section{Related Work}\label{Related Work}
iVQA is closely related to the tasks of image caption generation and visual question answering.
In this section, we provide a brief review over the related works and techniques for image caption generation,
visual question answering, and visual question generation.\\

\noindent \textbf{Image Caption Generation.}
 The standard CNN-RNN model \cite{r60} is considered the base for most state-of-the-art image captioning models. It uses a convolutional neural network (CNN) for visual feature extraction and uses recurrent neural networks (\eg LSTM) for caption generation. After the initial success of this simple model, Xu \etal~\cite{r51} improved it  by introducing a attention mechanism which helps the model to selectively focus on important parts of the image rather than consider its whole content. This powerful technique has inspired researchers to study different attention variants. 
You \etal \cite{r61} presented a semantic based attention strategy that helps the model to attend to the important semantic concepts 
of the input image. 
Lu \etal \cite{r62} proposed a more intuitive attention mechanism, namely adaptive attention. 
Their model enables the visual attention when there is a need to consider the visual cue 
and disables it when the visual cue does not correspond to the currently generated word. 
More recently, considerable improvements in image captioning have been 
achieved by combining both bottom-up and top-down attentions \cite{r44}.
The models developed for image captioning though are not directly applicable on other types of tasks
such as VQA, VQG, and iVQA, 
the principles of the techniques can be adapted and generalized.\\

\noindent\textbf{Visual Question Answering.}
VQA is similar to iVQA in the way that both tasks should capture complex textual and visual interactions 
and translate them into written words. 
Unlike image captioning, the primary challenge of VQA is to extract multi-modal joint features that sufficiently express 
relationships between the visual content and textual content.
To solve this problem, Zhou \etal \cite{r65} used the conventional concatenation method to combine textual and visual features. 
The moderate results of this straightforward method urged researchers to find other advanced techniques. 
One of such techniques is bilinear-pooling 
which allows both visual and textual features to fully interact with each other. 
However, the direct implementation of this method is computationally expensive 
and hence not 
applicable to the complex VQA task. 
Many algorithms were then developed to reduce the number of parameters needed to perform the bilinear-pooling operation. 
For example, the Multi-modal Low-rank Bilinear (MLB) pooling \cite{r59} 
decomposes the high dimensional weight tensor needed for the bilinear-pooling operation into three lower dimensional weight matrices which can easily fit into a GPU memory. 
Another method, 
Multi-modal Factorized Bilinear-pooling (MFB) \cite{r58} 
attempts to capture complex interactions by first expanding textual and visual features into a high-dimensional feature vector
and then 
%that is then 
squeezing it into a low-dimensional feature vector. 
MFB has demonstrated good performance for many state-of-the-art VQA models. 
In this work we also exploit MFB to capture complex interactions between answer and image cues.

Similar to image captioning, the attention mechanism also plays an important role in boosting VQA performance. Instead of considering the whole content of an image equally, the attention mechanism helps VQA models to focus on parts corresponding to the given question \cite{r66}. 
Many visual attention variants, e.g.,  
the Stacked attention \cite{r68} and Hierarchical co-attention \cite{r67}, 
have been proposed for VQA. 
Besides visual attention, semantic attention has also been explored; 
%%In this context, 
Yu \etal \cite{r64} designed a multi-level visual-semantic attention model 
that focuses on both important regions and semantic concepts corresponding to a given question. 
Nevertheless, as previously discussed, 
the rich semantic information contained in the questions could
make a VQA system produce the right answers without any 
image visual understanding \cite{r63},
which motivates the new task of iVQA.\\

\noindent\textbf{Visual Question Generation.}
VQG is a new deep learning task that aims to build an artificial intelligence system that can recognize the content of an image and generate relevant questions. One of the first research works that focused on this
task was conducted by Mostafazadeh \etal \cite{r6} who repurposed the standard CNN-RNN model for the task of VQG. 
Similar to that for image captioning, this model became the base for many advanced state-of-the-art VQA models. 
To enhance question diversity, %a more complex model
the model in \cite{r24} leveraged DenseCap \cite{r25} to generate questions that are corresponding to the main objects in a given image. 
%After that, 
Jain \etal \cite{r23} improved question diversity by integrating the vanilla variational autoencoder \cite{r35} into their model. 
Nevertheless, the aforementioned models generate questions merely based on images. 
Although this is very helpful for enhancing visual and linguistic understandings, 
the generated questions may be too generic and 
%not represent a
lack sufficient understanding of the given image. 
By contrast, the iVQA task proposed in \cite{r29} is more challenging % than VQG 
as an iVQA model requires precise understanding of a given image and 
good reasoning to generate specific questions that are corresponding to an image-answer pair. 
Two main approaches have been proposed for iVQA \cite{r29, r41}. 
In \cite{r29}, a powerful model is developed by incorporating two key components, 
the semantic concepts that are used together with a given answer to provide an initial first glance for question generation,
and the dynamic attention that dynamically focuses on important parts of a given image while generating the question word by word. 
In \cite{r41}, the authors leveraged a variational autoencoder to improve the capability of generating multiple %conditional 
corresponding questions. 
After developing %strong 
both VQA and VQG models, Li \etal \cite{r28} exploited the complementary relationship of the two tasks and proposed the first model that can work with both VQA and iVQA. This study constituted an important step to confirm the importance of iVQA in improving VQA.    

%-------------------------------------------------------------------------
\begin{figure}[t]
\begin{center}
\includegraphics[height=2.35in]{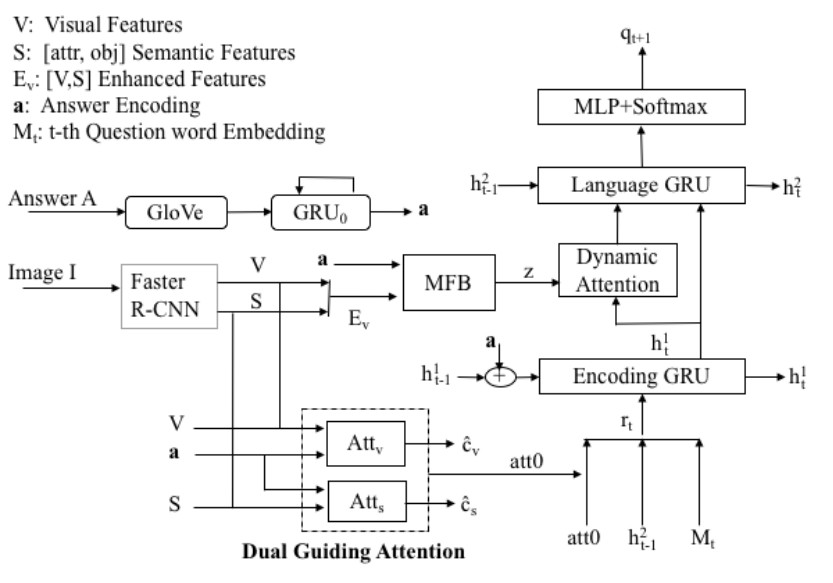}
\end{center}
\caption{The proposed multi-level attention iVQA model. 
The model consists of the following components:
(1) GRU-based answer encoder, which encodes the answer phrase into the semantic embedding space. 
(2) Faster R-CNN based image encoder, which produces visual and semantic features 
at the object and salient regions.
(3) Dual guiding attention, which produces answer attended visual and semantic features.
(4) Multi-modal feature fusion (MFB), which fuses the answer cue with the visual and semantic features.
(5) Two-layer GRU model, which recurrently generates the sequence of question words with a dynamic attention mechanism.
}
\label{fig:long}
\label{mainModel}
\label{fig:onecol}
\end{figure}

%-------------------------------------------------------------------------
\section{Approach}\label{Approach}   

Given an image $I$ and an answer phrase $A=[w_1,\cdots,w_m]$, 
inverse question answering (iVQA) generates a corresponding question 
$Q=[q_1,...,q_T]$ 
by maximizing the following conditional probability:
\begin{align}\label{iVQAmath}
Q^{\star} 
	&=\underset{Q}{\arg\max}\,P(Q|I, A;\Theta) \nonumber\\
	&=\underset{Q}{\arg\max} \prod_{t} P(q_t|q_1,\cdots, q_{t-1},I,A;\Theta)
\end{align}
where $\Theta$ denotes the model parameters. 
%-------------------------------------------------------------------------
\subsection{iVQA Model with Multi-Level Attentions}
In this section, we present a novel multi-level attention model for inverse question answering.
The architecture of the proposed model
is illustrated in Figure~\ref{mainModel}.

Comparing to previous iVQA methods, our proposed model has strengths in the following aspects:
(1) We encode the input image with 
a Faster R-CNN module, 
which generates both object level semantic labels (e.g., ``brown horse")
and associated regional visual features.
For iVQA, this image encoder 
enables consistent alignment of the visual image features
and textual answer cue,
while providing a suitable foundation for regional selective attention mechanisms.
(2) We use a dual guiding attention 
mechanism to attend the visual and semantic features based on the answer cue
for partial question encoding.
(3) We adopt the powerful 
Multi-modal Factorized Bilinear-pooling (MFB) \cite{r58} 
method to capture complex interactions between image and answer features
and fuse them into answer-aware image representations for dynamic attention
and question decoding (next word generation). 
%%%%
Below we present the key component modules of the proposed iVQA model.
%%%%

%-------------------------------------------------------------------------
   
\begin{figure}[t]
\begin{center}
\includegraphics[width=3.2in]{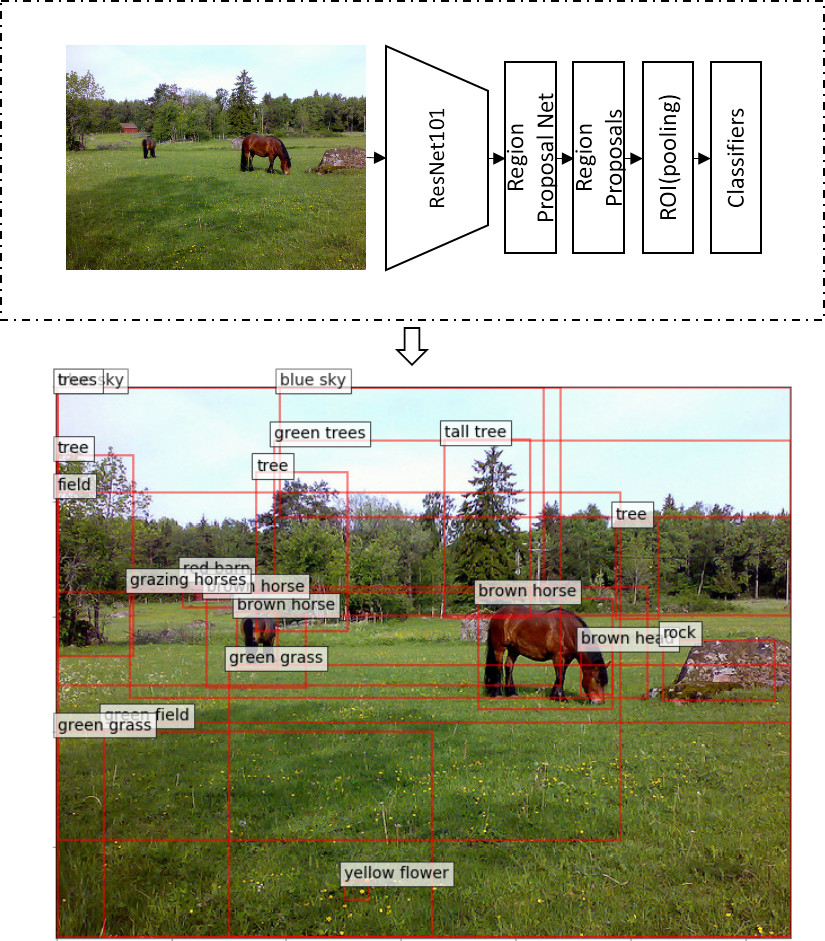}
\end{center}
\caption{
Illustration of the Faster R-CNN image encoder module.
The detected objects are illustrated in the bottom picture. 
Each object is localized with a bounding box and annotated with attribute and object labels.
	}
\label{fig:long}
\label{bottomUpModel}
\label{fig:onecol}
\end{figure}
%-------------------------------------------------------------------------
\subsubsection{Semantic Answer Encoder}
Comparing to questions, answers are typically very short phrases such that $A=[w_1,\cdots,w_m]$. 
Each word in the phrase has semantic meanings. 
To produce a semantically informative encoding for an answer phrase, 
we propose to exploit a natural language processing
technique, GloVe \cite{r56}, to encode
each word into a 300-dimensional semantic embedding vector.
Moreover, 
we use a recurrent neural network, Gated Recurrent Units (GRUs) \cite{r10}, 
to capture the dependencies between the sequence of words in an answer phrase
and use the hidden state output of the last GRU cell as the answer encoding vector,
${\bf a}\in\mathbb{R}^H$, with $H=1280$.

%-------------------------------------------------------------------------
\subsubsection{Semantic and Visual Image Encoder}
Different from previous iVQA approaches \cite{r29, r41}, which use visual features extracted from deep CNN models, \eg ResNet \cite{r2}, 
we adopt an object-based regional feature extraction model, 
Faster R-CNN \cite{r47} (with ResNet101 \cite{r2} pre-trained on ImageNet \cite{r48}),
to perform image encoding.
Faster R-CNN has been used as a state-of-the-art tool for object detection in computer vision. 
We use it to extract object-level visual features and associated object and attribute labels
from input images. The process is illustrated in Figure~\ref{bottomUpModel}.
This encoding module has two main steps. 
In the first step, the Faster R-CNN generates region proposals with their corresponding bounding boxes. 
Specifically we first pass a given image into a residual network (ResNet101) to extract a spatial feature map,
which is then passed into a Region Proposal Network (RPN) to generate object proposals 
with a sliding window.
For each window, there are multiple anchors
that consider different possibilities for a potential object location. 
At each anchor, RPN uses binary classification to predict the potential of
the current anchor forming an object-of-interest. 
Whenever RPN finds an object, it predicts a bounding box to localize this object. 
In the second step, the shapes of the region proposals are unified by applying a region-Of-Interest (ROI) pooling layer. 
The generated visual features are then passed to the final classification module,
which predicts objects and their corresponding attribute classes (\eg ``brown horse"),
while refining the corresponding boxes with regression.

The model is trained on the Visual Genome dataset \cite{r69}. 
For each image, we apply the encoder model to 
extract visual features from the top $k$  most probably predicted object regions 
and obtain final visual features,
$V=\{{\bf v}_i \in\mathbb{R}^{2048}|i= 1,\cdots,k\}$. 
To enhance these visual features 
with semantic information,
we use the predicted attribute label (e.g., ``brown") and object label (e.g., ``horse") on each of these regions. 
Similar to the answer encoder, we encode each label into a 300-dimensional embedding vector with GloVe \cite{r56}.
This leads to k regional attribute embeddings, $B=\{{\bf b}_i\in\mathbb{R}^{300}|i=1,\cdots,k\}$,
and object label embeddings, $O=\{{\bf o}_i\in\mathbb{R}^{300}|i=1,\cdots,k\}$.
Together they form the textual semantic features,
$S=\{{\bf s}_i=[{\bf b}_i;{\bf o}_i] |i=1,\cdots,k\}$,
in the same semantic space as the answer cue.
The visual and semantic features can also be concatenated to form
enhanced image representation $E_v$, such that
$E_v=\{E_{vi}=[{\bf v}_i; {\bf s}_i]|i=1,\cdots,k\}$.

%-------------------------------------------------------------------------
\subsubsection{Dual Guiding Attention}

Given the outputs of the answer encoder and image encoder, $\{{\bf a}, V, S\}$,
question generation is conducted by using a two-layer GRU model.
In order to provide the GRU model an initial idea about the 
importance of the semantic and visual concepts based on the answer cue
and guide the question generation,
we use a dual guiding attention mechanism to 
attend the visual and semantic representations in parallel. \\ 

\noindent\textbf{Visual Attention.}
The visual attention $Att_v$ aims to attend to the most important objects in the image based on the answer cue.
The attention weight for each regional visual vector, ${\bf v}_i \in V$, is computed as follows:
\begin{equation}
	\alpha_{i,v} = W_{att,v}^\top (\tanh(U_{v}{\bf v}_i+{\bf b}_{v}) \odot \tanh(U_{a}{\bf a}+{\bf b}_{a}))
\end{equation}
where $W_{att,v}\in \mathbb{R}^N$, $ U_{v}\in \mathbb{R}^{N\times N_v}$, ${\bf b}_v, {\bf b}_a\in\mathbb{R}^N$, 
and $U_{a}\in \mathbb{R}^{N\times H}$ 
are the model parameters, 
and $\odot$ denotes the Hadamard product operator.
We then pass the visual attention weights into a softmax layer to compute the 
attended visual representation of the image, ${\bf \hat{c}}_v$:
\begin{align}
	{\bf \beta}_v = \mbox{softmax}({\bf \alpha}_v),
	\qquad
	{\bf \hat{c}}_v = \sum_{i=1}^{i=k}\beta_{i,v}{\bf v}_i
\end{align}

\noindent\textbf{Semantic Attention.}
We compute the attended semantic representation of the image in a similar way. 
The attention weight for 
each semantic feature vector, ${\bf s}_i\in S$, is computed as follows:
\begin{equation}
	\alpha_{i,s} = W_{att,s}^\top (\tanh(U_{s}{\bf s}_i+{\bf b}_{s}) \odot \tanh(U'_{a}{\bf a}+{\bf b'}_{a}))
\end{equation}
where $W_{att,s}\in \mathbb{R}^{N}$, 
$U_{s}\in \mathbb{R}^{N\times 600}$, ${\bf b}_s, {\bf b'}_a\in\mathbb{R}^N$, 
and $U'_{a}\in \mathbb{R}^{N\times H}$ 
are the model parameters. 
The attended semantic features,
${\bf \hat{c}}_s$, is then computed as:
\begin{align}
	{\bf \beta}_s = \mbox{softmax}({\bf \alpha}_s),
	\qquad
	{\bf \hat{c}}_s = \sum_{i=1}^{i=k}\beta_{i,s}{\bf s}_i
\end{align}
The attended visual and semantic features are then concatenated to form the input features, 
$att_0=[{\bf \hat{c}}_v;{\bf \hat{c}}_s]$, for the partial question encoding GRU.

%-------------------------------------------------------------------------
\subsubsection{Multi-Modal Feature Fusion}
To capture complex interactions between the image and answer cue,
we use a multi-modal factorized bilinear pooling (MFB) operation \cite{r58} 
to fuse the image and answer representation vectors. 
MFB is developed 
to reduce the computational cost of the standard bilinear-pooling operation between question and visual features
for visual question answering. 
We adopt MFB to fuse the answer encoding vector ${\bf a}$ and 
the enhanced image representation $E_v$.

MFB is a two-step algorithm with the expansion and squeeze steps.
For each enhanced visual feature vector
$E_{vi}$ and the answer encoding vector ${\bf a}$,
we first project them into high dimensional space to fuse 
and then squeeze the fused vector using
a sum pooling operation with window size $j$:
\begin{align}
	{\bf f}_i &= (U_1E_{vi} + {\bf b}_1) \odot (U_2{\bf a} +{\bf b}_2),
	\\
	{\bf z}_i &= \mbox{sum-pooling}({\bf f}_i, j)
\end{align}
where $U_1\in \mathbb{R}^{N_f\times N_e}$ and $U_1\in \mathbb{R}^{N_f\times H}$ 
are the high dimensional projection weight matrices,
${\bf b}_1$ and ${\bf b}_2$ are the bias parameter vectors,
$j$ is the MFB parameter ($\eg$ 5). 
The MFB feature vector ${\bf z}_i$ is then processed by a signed square root normalization and $L_2$ normalization.

%-------------------------------------------------------------------------
\subsubsection{Recurrent Question Generation}

Given the concatenation of the attended visual and semantic vectors, $att_0$, 
and the fused image and answer representation $\{{\bf z}_i\}$, 
we deploy a two-layer Gated recurrent unit (GRU) model 
for question generation, 
which generates the sequence of question words, $[q_1,\cdots, q_T]$, recurrently.
This module has three main components for each recurrent generation operation:
partial question encoder, dynamic attention, and question decoder. \\

%--------------------------------------------------------------------------
\noindent\textbf{Partial Question Encoder.}
At each recurrent step, a GRU encoder is deployed to encode
the current partial question generated, aiming to capture 
the long-term dependencies in the question word sequence for next word generation.
Given the question word $q_t$ generated from the immediate previous recurrent step,
we first embed
$q_t$ into the same semantic embedding space 
as the answer cue 
with a pre-trained GloVe model \cite{r56},
which produces an embedding vector $M_t\in\mathbb{R}^{300}$.
Specifically, we perform the word embedding as follows: 
\begin{equation}
	M_t = W_{emb} {\bf 1}_{q_t} + {\bf b}_{emb}
\end{equation}
where $W_{emb}\in \mathbb{R}^{300\times N_w}$ is initialized from the word embedding matrix produced by GloVe,
$N_w$ denotes the vocabulary size, 
${\bf b}_{emb}$ is the bias vector,
and ${\bf 1}_{q_t}$ denotes an one-hot vector with length $N_w$, which has a single $1$ for the entry corresponding to word $q_t$.

Then we update the recurrent question encoding state ${\bf h}^1_t$ by using the concatenation
of the word embedding $M_t$, answer attended visual and semantic features $Att_0$,
and the previous question decoding state ${\bf h}^2_{t-1}$ as input:
\begin{align}
	{\bf r}_t &= [M_t; {\bf h}^{2}_{t-1}; att_0]
	\\
	{\bf h}^1_t&=\mbox{GRU}_1({\bf r}_t, {\bf h}^1_{t-1} + {\bf a})
\end{align} 
Note, we further emphasize the answer cue by adding it to 
the previous hidden state vector of the encoder GRU$_1$. \\

%--------------------------------------------------------------------------
\noindent\textbf{Dynamic Attention.}
Given the current partial question encoding state, we use an attention module
to dynamically assign higher weights 
to important regional visual features that are corresponding to current partial question context. 
Specifically, the attention weights are determined by a triplet of image, answer, and partial question encoder state.
We compute them as follows:
\begin{equation}
	\label{dattw}
	\alpha_{i,t} = W_{att}^\top\tanh(W_{z}{\bf z}_i+{\bf b}_{z} + W_{h}{\bf h}^1_t)
\end{equation}
where $W_{att}\in \mathbb{R}^{N}$, $W_z\in \mathbb{R}^{N\times N_z}$, 
${\bf b}_z\in\mathbb{R}^N$ and $W_h\in \mathbb{R}^{N\times N_h}$ 
are the model parameters. 
The attention weights are further normalized by a softmax function
to compute the dynamically attended image visual features:
\begin{align}
	{\bf \beta}_{t}= \mbox{softmax}({\bf \alpha}_{t}),
	\qquad
	{\bf\hat{c}}_t = \sum_{i=1}^{i=k}\beta_{i,t}{\bf v}_i
\end{align}
Note that the attention weights are determined by two factors, $\{{\bf z}_i\}$ and ${\bf h}^1_t$. 
The former  $\{{\bf z}_i\}$  fused the answer cue into each regional image representation vector with the MFB operation,
while the latter dynamically reflects the current partial question context.\\

%--------------------------------------------------------------------------
\noindent\textbf{Question Decoder.}
In the top GRU decoding layer, which is referred to as language GRU, 
we concatenate the dynamically attended image visual feature vector ${\bf\hat{c}}_t$  
with the output state of the GRU encoder ${\bf h}^1_t$
to update the state of the recurrent GRU decoder:
\begin{equation}
	\mbox{h}^2_t = \mbox{GRU}_2([{\bf \hat{c}}_t, {\bf h}_t^1], {\bf h}^2_{t-1})
\end{equation}
where ${\bf h}^2_{t-1}$ is the previous hidden state at step $t-1$ and it is set to zero when $t=0$.

To predict the next word probability distribution over the set of vocabulary words, 
we further pass the state vector of the GRU decoder through a fully connected layer and a softmax layer: 
\begin{equation}\label{finalProb}
	{\bf p}^* =  \mbox{softmax}(W_{dec}{\bf h}^2_t + {\bf b}_{dec})
\end{equation}
where $W_{dec}\in \mathbb{R}^{N_w\times N_h}$ and ${\bf b}_{dec}\in\mathbb{R}^{N_w}$ are the model parameters.
With beam size =1, the next $(t+1)$-th word can then be generated as:
\begin{equation}\label{finalword}
	q_{t+1} = \mbox{VOC}(j^*),\qquad
	j^* = {\arg\max}_{j}\; {\bf p}_j^*
\end{equation}
where VOC denotes the vocabulary table.

%-------------------------------------------------------------------------
\subsection{Training and Inference}
We perform an end-to-end deep training over a set of training instances.
The $i$-th instance is a triplet that includes 
an image $I^i$, 
an answer phrase $A^i=[w^i_1,\cdots,w^i_{m_i}]$, 
and a ground truth question  
$Q^i=[q^i_1,...,q^i_{T_i}]$.
We train the proposed model 
by minimizing the cross-entropy loss:
\begin{equation}
	L(\Theta) = -\sum_i \frac{1}{T_i}\sum_{t=1}^{t=T_i} \log P(q^i_t|q^i_1,\cdots,q^i_{t-1};I^i,A^i)
\end{equation}
where $\Theta$ represents all the model parameters, 
and $P(q^i_t|q^i_1,\cdots,q^i_{t-1};I^i,A^i)$ is determined from Eq.(\ref{finalProb}).

During the test phase, given the trained model and an image-answer pair,
we recurrently predict 
the probability distribution of the next word in the question sequence over our vocabulary
using Eq.(\ref{finalProb}).
We use a beam size 1 (greedy search) to select the most probable word as the output,
as in  Eq.(\ref{finalword}). 
The question generation terminates after generating the question mark.

%-------------------------------------------------------------------------

\begin{table*}[th!]
\centering
\setlength{\tabcolsep}{10pt}
\caption{
Comparison results between our proposed iVQA model 
and the other three comparison models on the popular Karpathy test split. 
The numbers in bold denote the highest scores.
}
\begin{tabular}{ |l|c|c|c|c|c|c|c| } 
\hline
Method & BLEU-1 & BLEU-2 & BLEU-3 & BLEU-4 & METEOR & ROUGE-L & CIDEr\\
\hline 
SAT \cite{r29} & 0.417 & 0.311 & 0.241 & 0.192 & 0.195 & 0.456 & 1.533 \\
iVQA+VAE \cite{r41} & 0.421 & 0.320 & 0.253 & 0.205 & 0.201 & 0.466 & 1.682 \\ 
iVQA \cite{r29} & 0.430 & 0.326 & 0.256 & \textbf{0.208} & 0.205 & 0.468 & \textbf{1.714} \\ 
\hline
Ours & \textbf{0.452} & \textbf{0.339} & \textbf{0.262} & 0.207 & \textbf{0.209} & \textbf{0.472} & 1.673 \\
\hline
\end{tabular}
\label{testResult}
\end{table*}

%-------------------------------------------------------------------------
\section{Experiments}\label{Experiments}
In this section we present our empirical study and report both quantitative and qualitative results.

\subsection{Experimental Setting}
\noindent\textbf{Dataset.}
We used the VQA V1 dataset \cite{r17} to conduct experiments and evaluate the proposed approach. 
This dataset was initially created for visual question answering 
and repurposed for iVQA \cite{r29} later.
The images are collected from the MS COCO dataset \cite{r45}.
The dataset is divided into three parts: training set, validation set and testing set. 
The training set consists of 82,783 images. The validation and testing sets include 40,504 and 81,434 images respectively. 
The answers however are not available for the testing set.
Therefore, we followed \cite{r29,r44,r46} to use the Karpathy data split,
and get
5000 images for testing and 5000 images for validation.
Following \cite{r29}, we use three question-answer pairs for each image.
This results in about 250k instances ((image, answer, question)-triplets) for training
and 15k instances for testing.\\

%-------------------------------------------------------------------------
\noindent\textbf{Implementation Details.}
Before conducting experiments, 
we first performed pre-processing over the data.
We follow \cite{r70} to extract the most frequent 3,000 answers from the training set 
and then build the vocabulary from the questions corresponding to these answers. 
This leads to a vocabulary set of about 12,900 words. 
The answers and questions are tokenized 
into lists of words. 
We standardized the lengths of questions and answers to be 19 and 3 respectively 
by padding zeros to shorter sequences and trimming longer sequences.
To extract the visual and semantic information, 
we use the publicly available bottom-up attention model, Faster R-CNN, in \cite{r44}. 
We use the model to extract the top $k=36$ salient regions of visual features $V$ and their corresponding semantic information $S$.

We performed training by using one GPU with 16 GB of memory. 
We fed the data in batches of 1000 instances. 
We use $N_h=1280$ hidden units for each GRU cell and $N=512$ for each attention module. 
We trained our model by using the back-propagation algorithm with the Adam \cite{r49} optimizer. 
During training, we initially set the learning rate as $9.9e^{-4}$. 
After the fifth epoch, we decreased it to $9.9e^{-5}$ 
and stopped training at the 14th epoch. \\

%-------------------------------------------------------------------------
\noindent\textbf{Model Evaluation.}
We evaluate our model by using the standard evaluation metrics, 
specifically, BLEU, METEOR, ROUGE-L and CIDEr%
\footnote{\url{https://github.com/tylin/coco-caption}}. 
These metrics are used extensively for image captioning evaluation \cite{r45} 
and have also been used for iVQA evaluation~\cite{r29}.

%-------------------------------------------------------------------------
\subsection{Quantitative Results}
To validate the effectiveness of the proposed model, 
we compare its performance with one baseline and two iVQA competitors \cite{r29, r41}. 
The baseline is the powerful Show Attend and Tell (SAT) model \cite{r30}, 
which was mainly developed for the task of image captioning and modified by \cite{r29} to take a given answer cue. 
The model is compared to ours in the context of using dynamic attention to focus on different salient regions at each time step. 
The competitors are the original iVQA method \cite{r29} and its modified version (iVQA+VAE) \cite{r41} 
which incorporates the variational autoencoder (VAE) to increase the question diversity. 
We report the comparison results in Table \ref{testResult}.

From the comparison scores, 
we can see that the proposed model with the multi-level attentions and semantically enhanced visual features 
performs better than the baseline SAT by remarkable margins. 
This suggests that the iVQA task presents different challenges than the caption generation task, 
while a state-of-the-art image captioning model can fail to produce effective results on the iVQA task
even after adaptation.
Moreover, we can also see that the proposed model also 
outperforms the state-of-the-art iVQA models, iVQA and iVQA+VAE, 
in terms of most of the linguistic evaluation metrics,
including BLEU-1, BLEU-2, BLEU-3, METEROR, and ROUGE-L. 
The results are also achieved under an inference setting that is favorable to the competitors
--
the two competitors used a beam search of size 3, while we only used a beam search of size 1. 
These impressive results demonstrate the efficacy of the proposed multi-level attention model.

\begin{table*}[th!]
\centering
\setlength{\tabcolsep}{9pt}
\caption{Performance comparison of our main model variants.}
\begin{tabular}{ |l|c|c|c|c|c|c|c| } 
\hline
Method & BLEU-1 & BLEU-2 & BLEU-3 & BLEU-4 & METEOR & ROUGE-L & CIDEr\\
\hline 
Variant w/o \{S, att\_0\}  & 0.450 & 0.336 & 0.259 & 0.204 & 0.208 & 0.467 & 1.649 \\
Full Model & {0.452} & {0.339} & {0.262} & 0.207 & {0.209} & {0.472} & 1.673 \\
\hline
\end{tabular}
\label{ablationStudy}
\label{testResult}
\end{table*}

%-------------------------------------------------------------------------
\subsection{Ablation Study} 
The proposed model incorporates additional semantic information $S$
based on the object and attribute labels. 
It uses $S$ to produce enhanced visual features $E_v$ 
and obtain $att0$ through dual guiding attention.
To investigate whether and how much this semantic information contributes to the major performance gain 
for the proposed model, we create a variant of the proposed model 
by dropping the semantic information $S$ and dual guiding attention features $att0$.
We compared this variant with the full model and reported the results in 
Table \ref{ablationStudy}.
We can see that the performance of the variant model 
is comparable to the state-of-the-art iVQA model \cite{r29}. 
It actually outperforms the iVQA model in terms of BLEU-1, BLEU-2, BLEU-3 and METEOR.
This validates 
the visual features extracted at the object-level salient regions more suitable for the task of iVQA
than the visual features extracted with conventional deep networks. 
Nevertheless, the full model still beats the variant in terms of all evaluation metrics.
This suggests that the use of semantic information and dual guiding attention is beneficial for iVQA. 
Enhancing visual features with semantic information that has the same textual modality with the answer 
and generated partial question is useful for our model to precisely attend to corresponding objects, 
and consequently improve the overall performance.
%------------------------------------------------------------------------- 
%------------------------------------------------------------------------- 
\begin{figure*}
\captionsetup[subfigure]{labelformat=empty}  
    \begin{subfigure} [b]{0.160\textwidth}        
            \includegraphics[width=1.0in,height=0.88in]{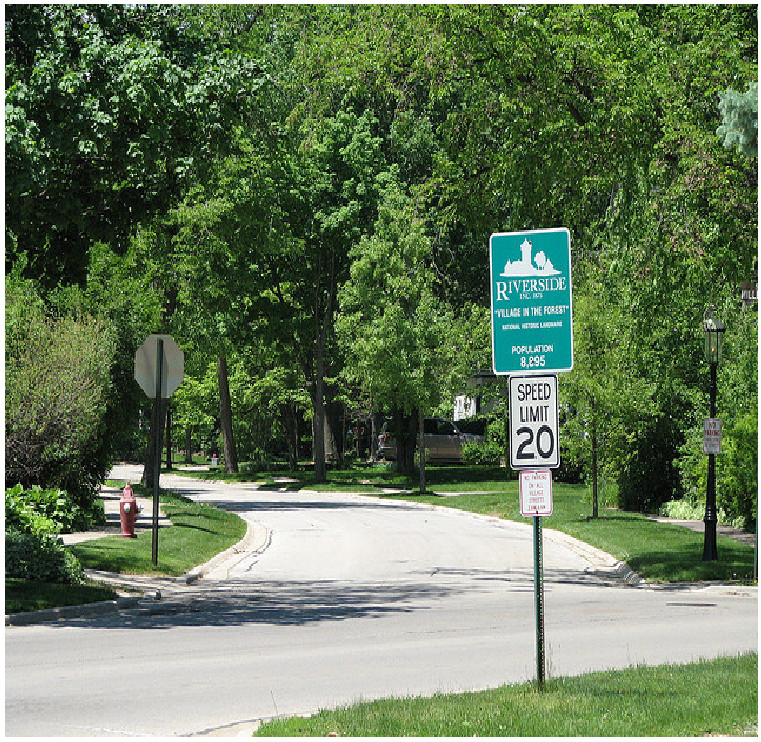}
            \caption{(a) Answer: green}
            \label{fig:SRl}
    \end{subfigure}%
     \hspace{0.015\textwidth}
    \begin{subfigure} [b]{0.160\textwidth}     
            \includegraphics[width=1.0in,height=0.88in]{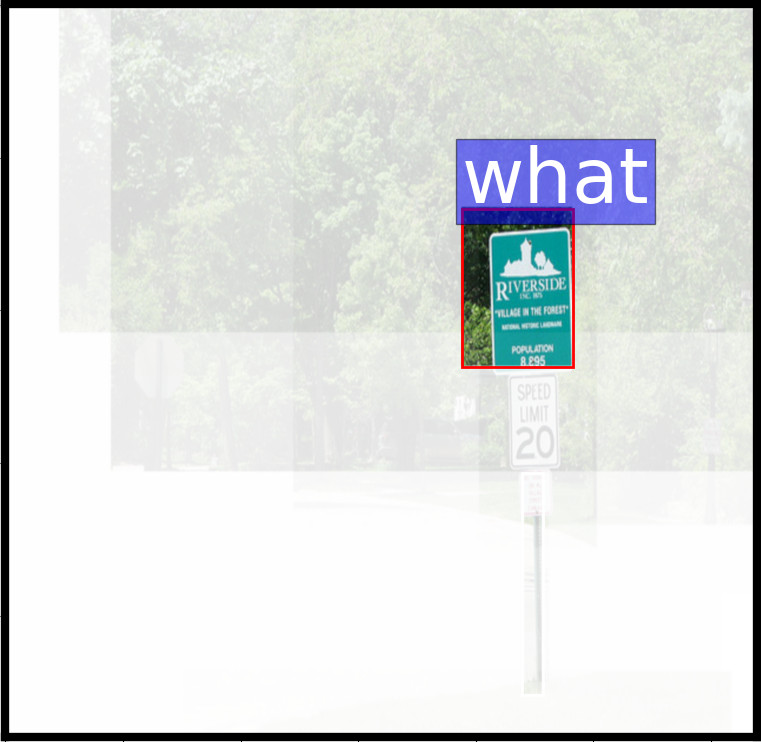}
            \caption{What}
            \label{fig:SRl}
    \end{subfigure}%
    \begin{subfigure} [b]{0.160\textwidth}        
            \includegraphics[width=1.0in,height=0.88in]{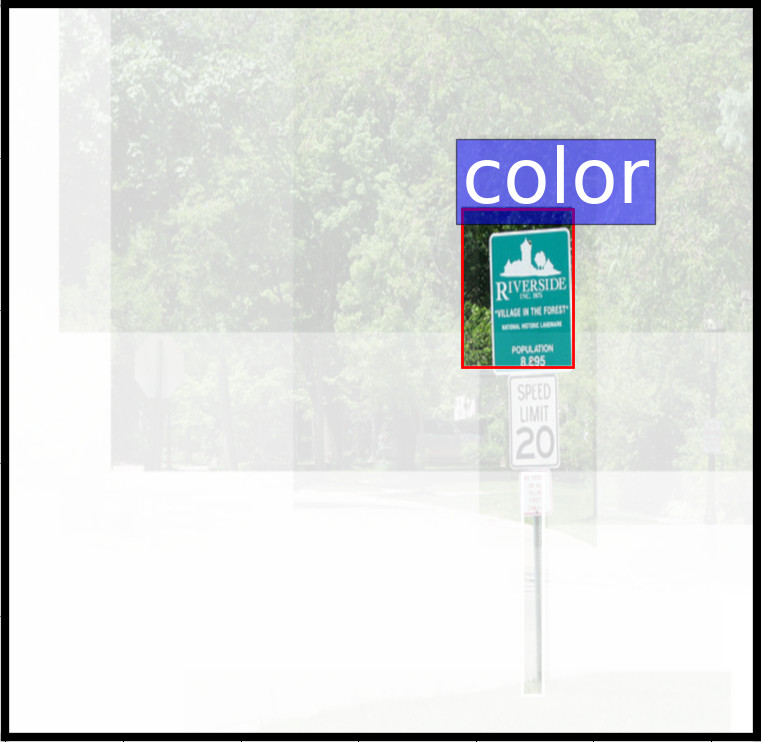}
            \caption{color}
            \label{fig:SRl}
    \end{subfigure}%
    \begin{subfigure} [b]{0.160\textwidth}        
            \includegraphics[width=1.0in,height=0.88in]{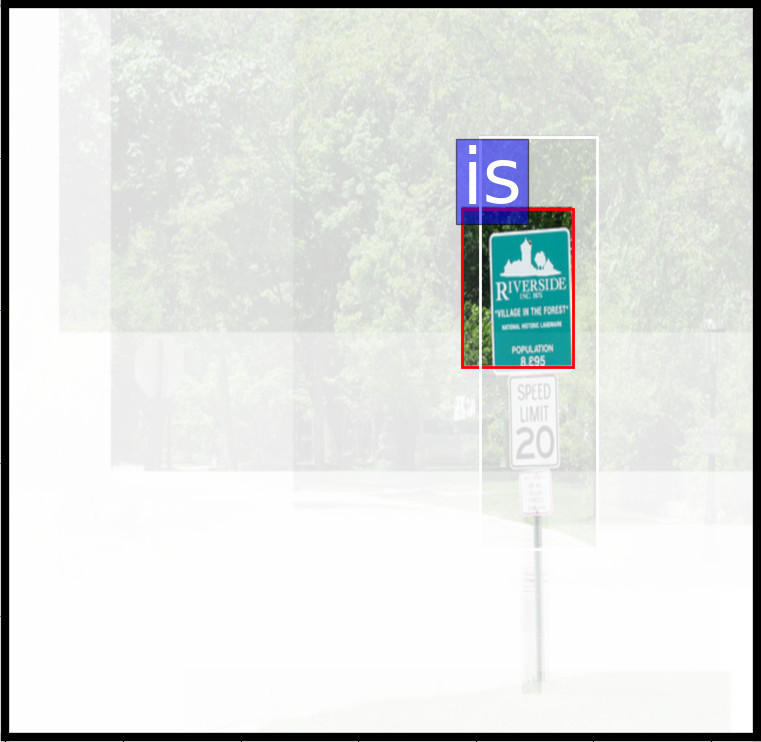}
            \caption{is}
            \label{fig:SRl}
    \end{subfigure}%
    \begin{subfigure} [b]{0.160\textwidth}        
            \includegraphics[width=1.0in,height=0.88in]{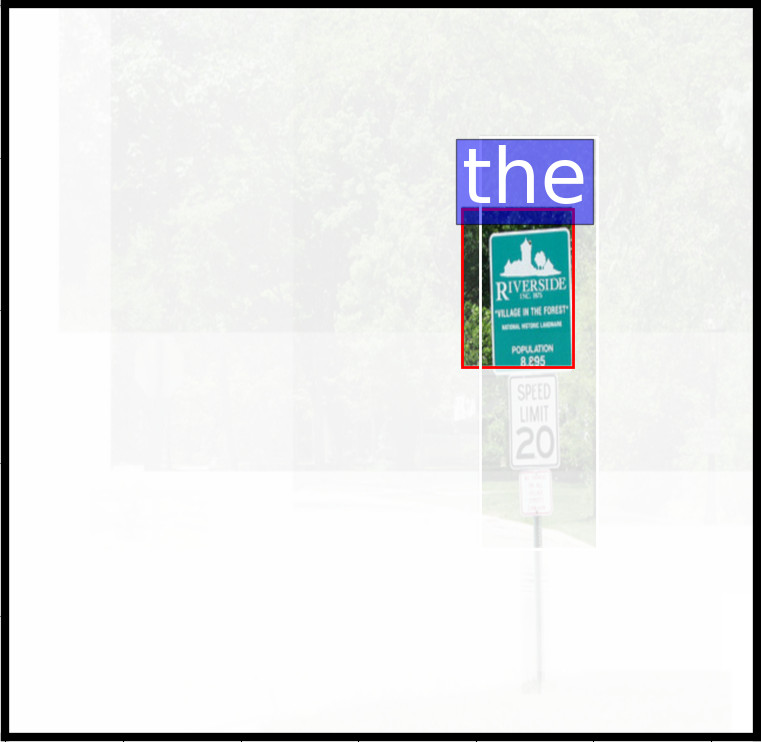}
            \caption{the}
            \label{fig:SRl}
    \end{subfigure}%
    \begin{subfigure} [b]{0.160\textwidth}        
            \includegraphics[width=1.0in,height=0.88in]{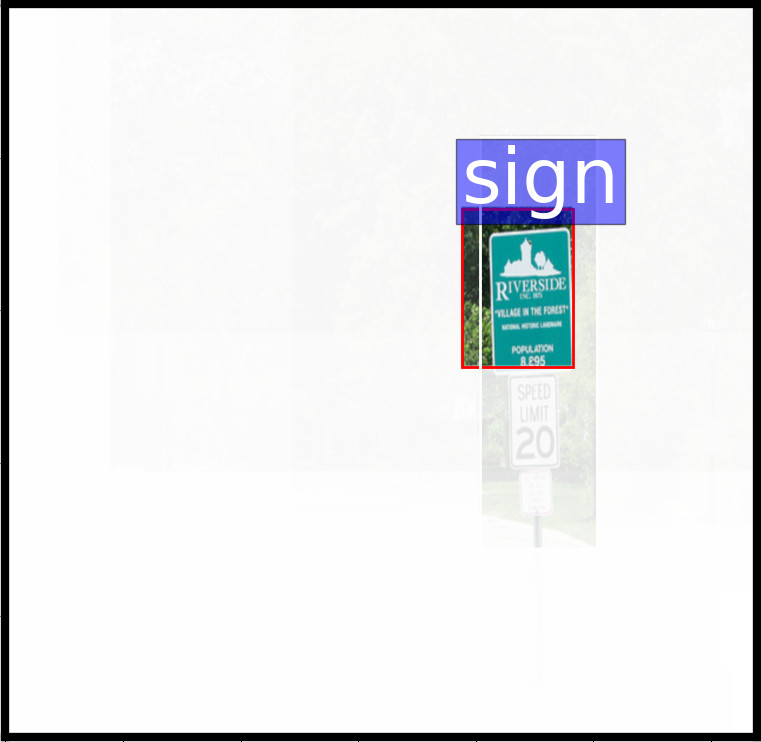}
            \caption{sign}
            \label{fig:SRl}
    \end{subfigure}%
    \\
    %%%%%%%%%%%%%%%%%%%%%%%%%%%%%%%%%%%%%%%%%%%%%%%%%%%%%%%%%%%%%%%%%%%
    \begin{subfigure} [b]{0.160\textwidth}        
            \includegraphics[width=1.0in,height=0.88in]{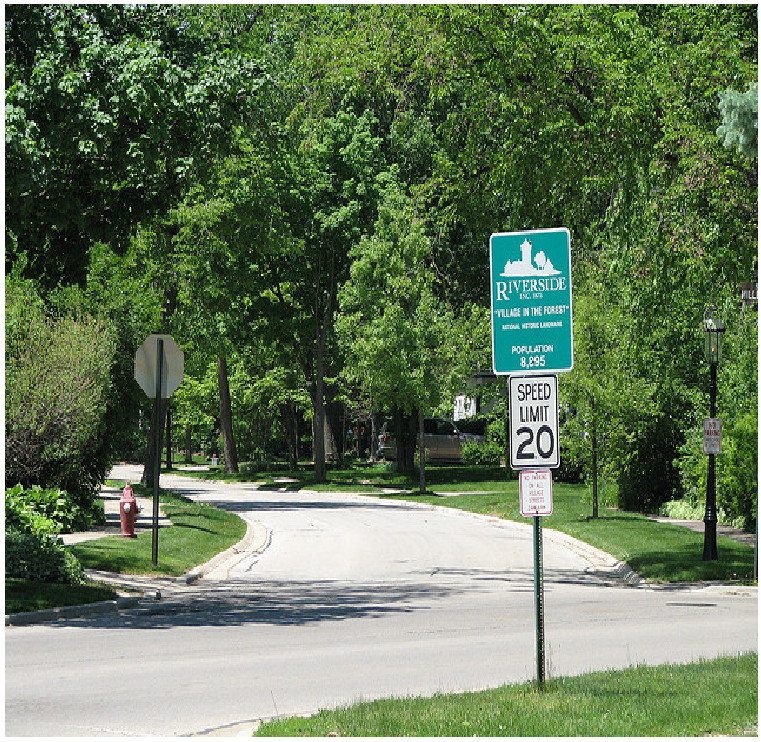}
            \caption{(b) Answer: red}
            \label{fig:SRl}
    \end{subfigure}%
     \hspace{0.015\textwidth}
    \begin{subfigure} [b]{0.160\textwidth}     
            \includegraphics[width=1.0in,height=0.88in]{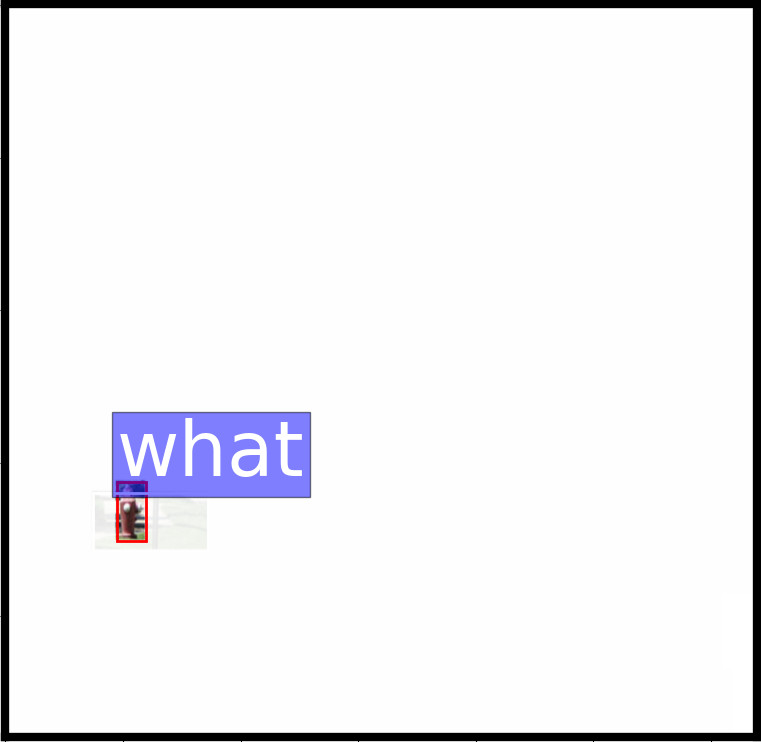}
            \caption{What}
            \label{fig:SRl}
    \end{subfigure}%
    \begin{subfigure} [b]{0.160\textwidth}        
            \includegraphics[width=1.0in,height=0.88in]{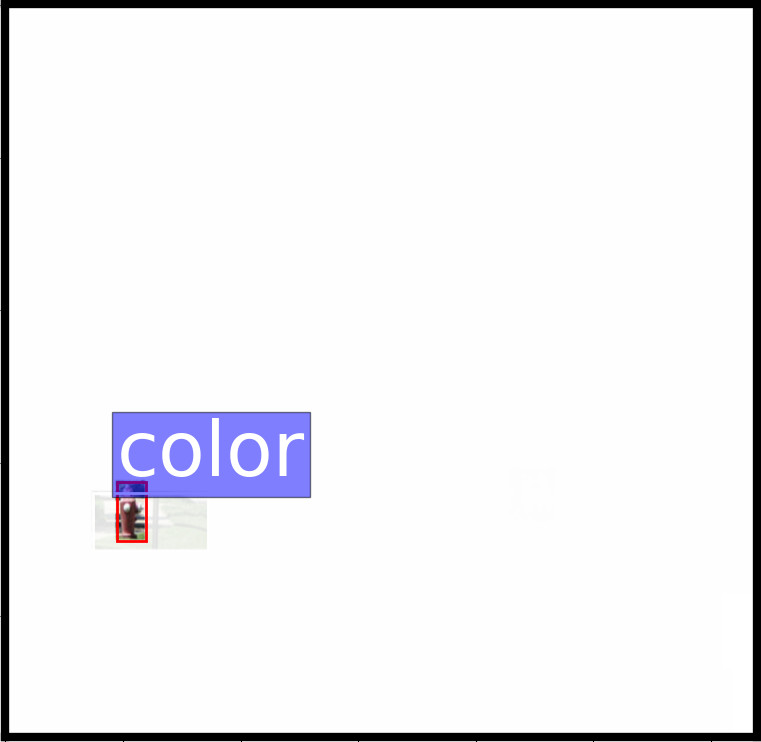}
            \caption{color}
            \label{fig:SRl}
    \end{subfigure}%
    \begin{subfigure} [b]{0.160\textwidth}        
            \includegraphics[width=1.0in,height=0.88in]{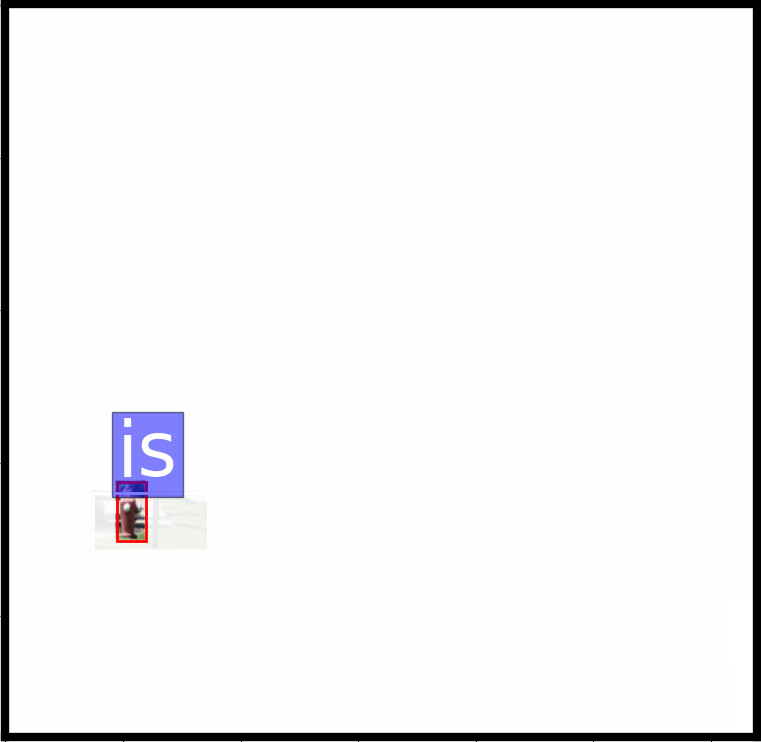}
            \caption{is}
            \label{fig:SRl}
    \end{subfigure}%
    \begin{subfigure} [b]{0.160\textwidth}        
            \includegraphics[width=1.0in,height=0.88in]{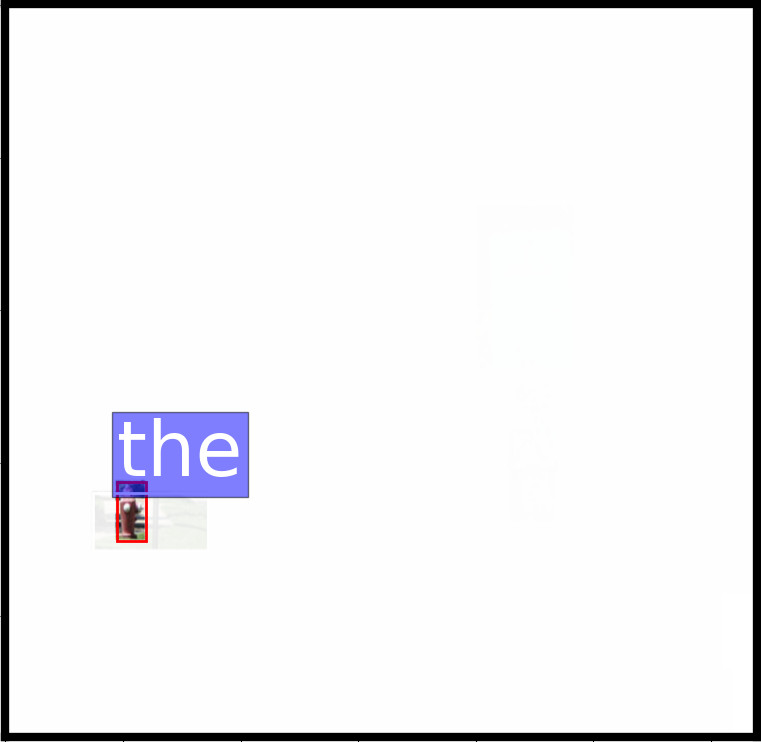}
            \caption{the}
            \label{fig:SRl}
    \end{subfigure}%
    \begin{subfigure} [b]{0.160\textwidth}        
            \includegraphics[width=1.0in,height=0.88in]{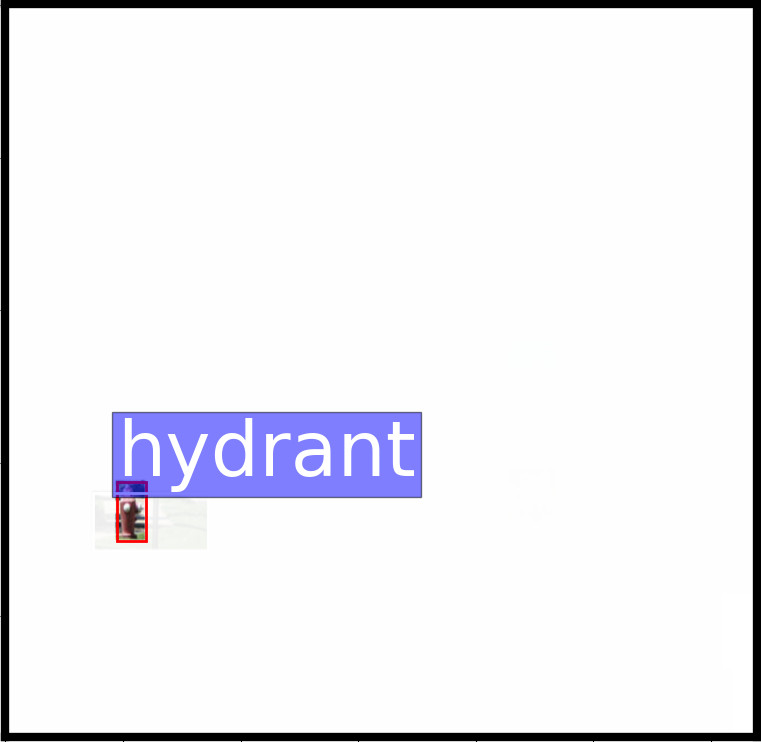}
            \caption{hydrant}
            \label{fig:SRl}
    \end{subfigure}%
    \\
    %%%%%%%%%%%%%%%%%%%%%%%%%%%%%%%%%%%%%%%%%%%%%%%%%%%%%%%%%%%%%%%%%%
    \begin{subfigure} [b]{0.160\textwidth}        
            \includegraphics[width=0.88\textwidth]{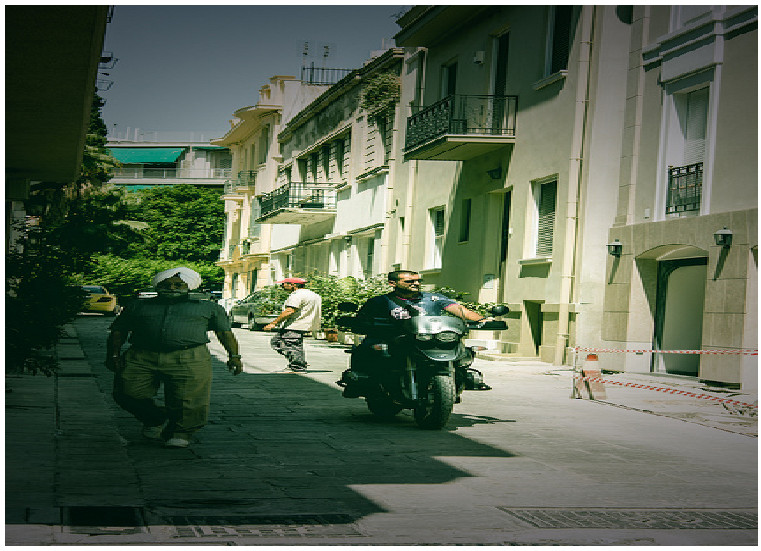}
            \caption{(c) Answer: turban}
            \label{fig:SRl}
    \end{subfigure}%
    \hspace{0.015\textwidth}
    \begin{subfigure} [b]{0.160\textwidth}     
            \includegraphics[width=0.88\textwidth]{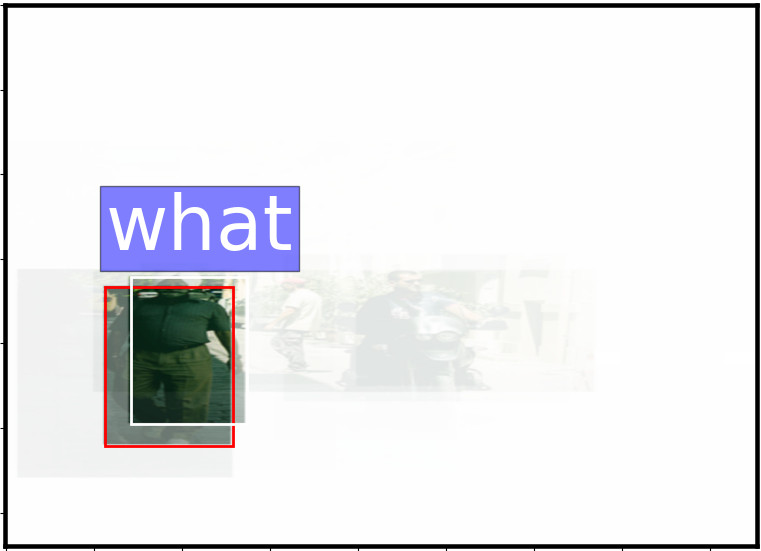}
            \caption{What}
            \label{fig:SRl}
    \end{subfigure}%
    \begin{subfigure} [b]{0.160\textwidth}        
            \includegraphics[width=0.88\textwidth]{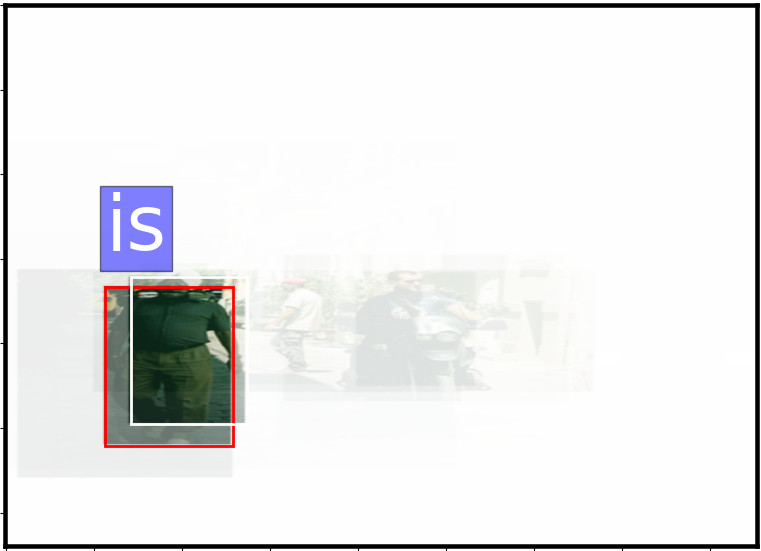}
            \caption{is}
            \label{fig:SRl}
    \end{subfigure}%
    \begin{subfigure} [b]{0.160\textwidth}        
            \includegraphics[width=0.88\textwidth]{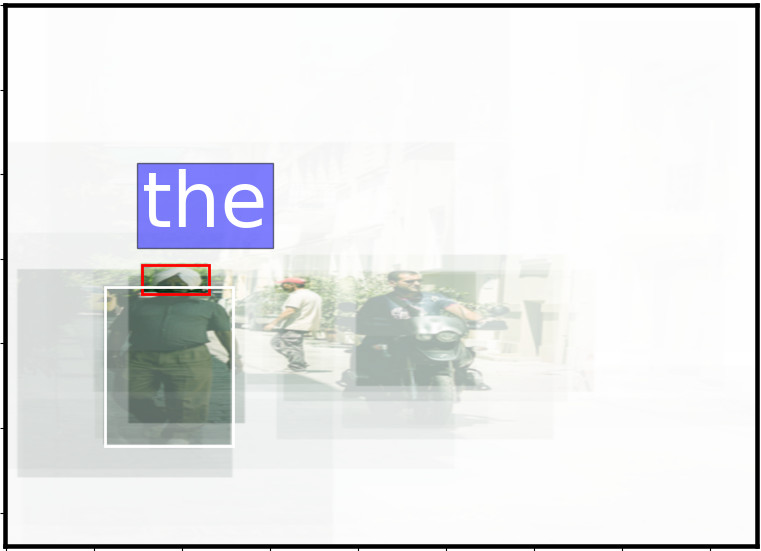}
            \caption{the}
            \label{fig:SRl}
    \end{subfigure}%
    \begin{subfigure} [b]{0.160\textwidth}        
            \includegraphics[width=0.88\textwidth]{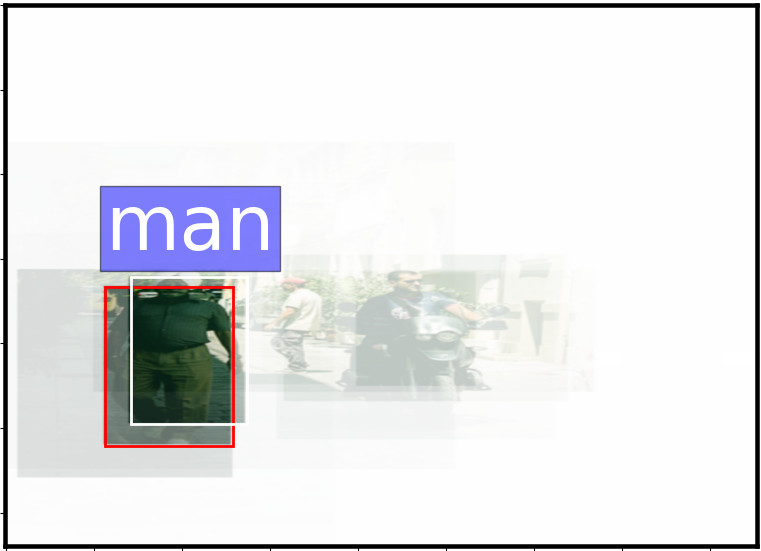}
            \caption{man}
            \label{fig:SRl}
    \end{subfigure}%
    \begin{subfigure} [b]{0.160\textwidth}        
            \includegraphics[width=0.88\textwidth]{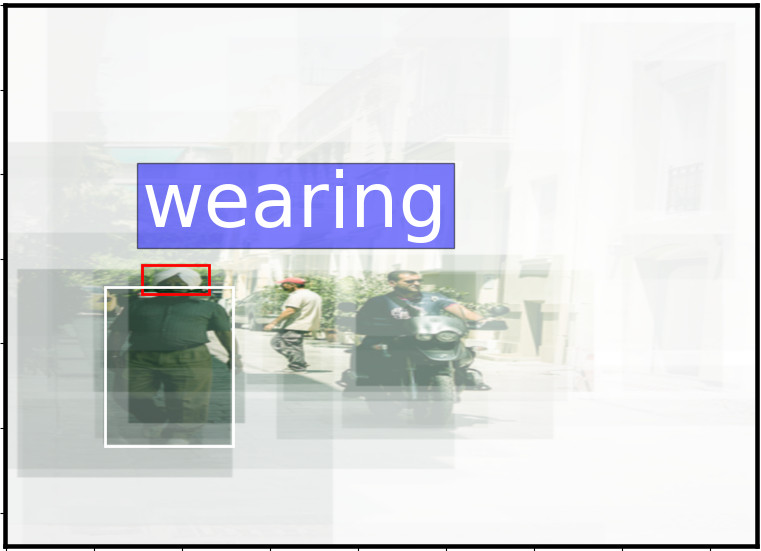}
            \caption{wearing}
            \label{fig:SRl}
    \end{subfigure}%
    \\ 
    \begin{subfigure} [b]{0.160\textwidth}            
    \end{subfigure}%
    \hspace{0.181\textwidth}%
    \begin{subfigure} [b]{0.160\textwidth}        
            \includegraphics[width=0.88\textwidth]{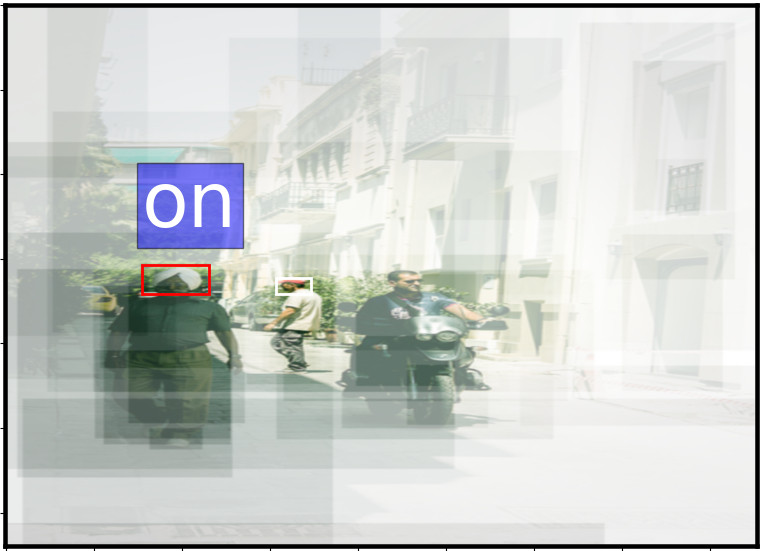}
            \caption{on}
            \label{fig:SRl}
    \end{subfigure}%
    \begin{subfigure} [b]{0.160\textwidth}        
            \includegraphics[width=0.88\textwidth]{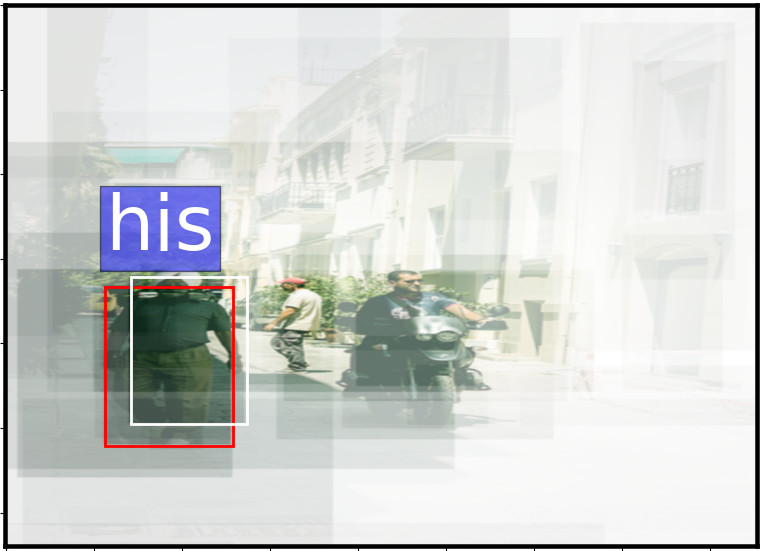}
            \caption{his}
            \label{fig:SRl}
    \end{subfigure}%
    \begin{subfigure} [b]{0.160\textwidth}        
            \includegraphics[width=0.88\textwidth]{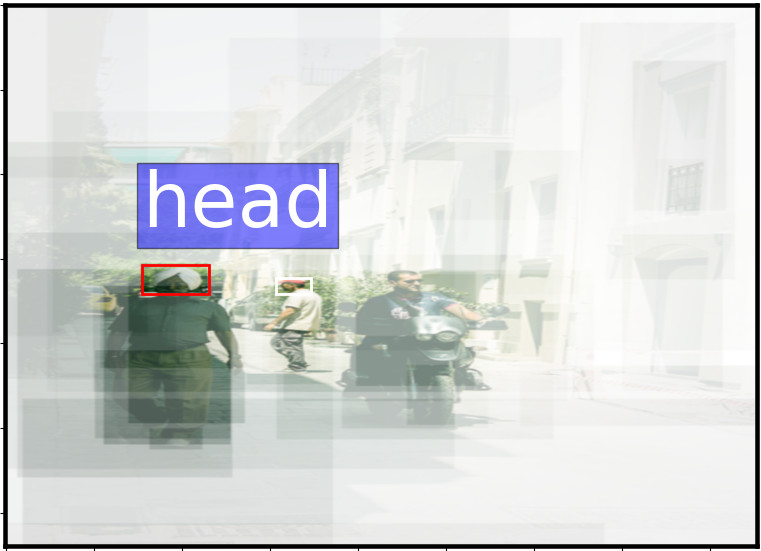}
            \caption{head}
            \label{fig:SRl}
    \end{subfigure}%
    \\
%%%%%%%%%%%%%%%%%%%%%%%%%%%%%%%%%%%%%%%%%%%%%%%%%%%%%%%%%%%%%%%%%%%
	\begin{subfigure} [b]{0.160\textwidth}        
            \includegraphics[width=1.0in,height=0.88in]{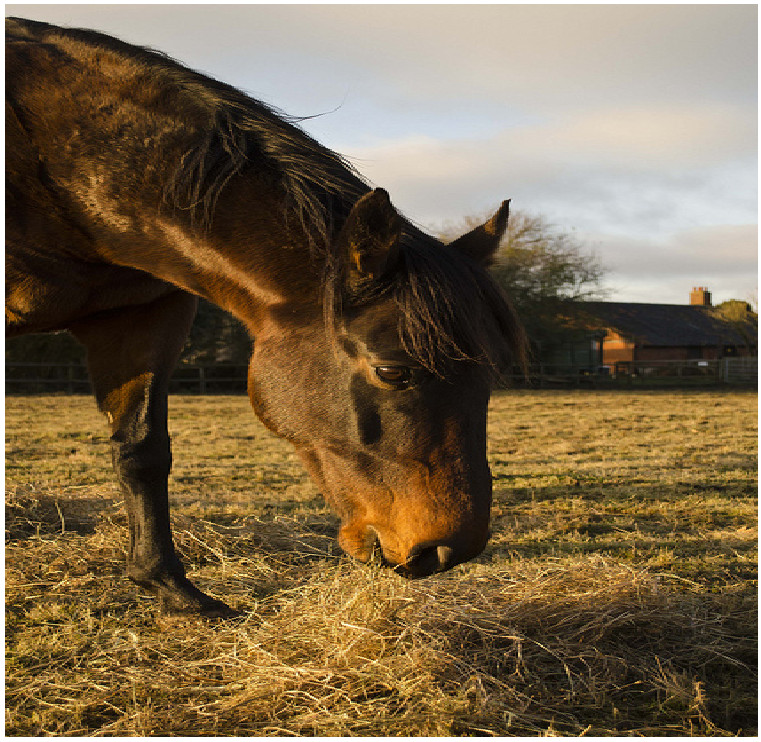}
            \caption{(d) Answer: hay}
            \label{fig:SRl}
    \end{subfigure}%
    \hspace{0.015\textwidth}    
    \begin{subfigure} [b]{0.160\textwidth}        
            \includegraphics[width=1.0in,height=0.88in]{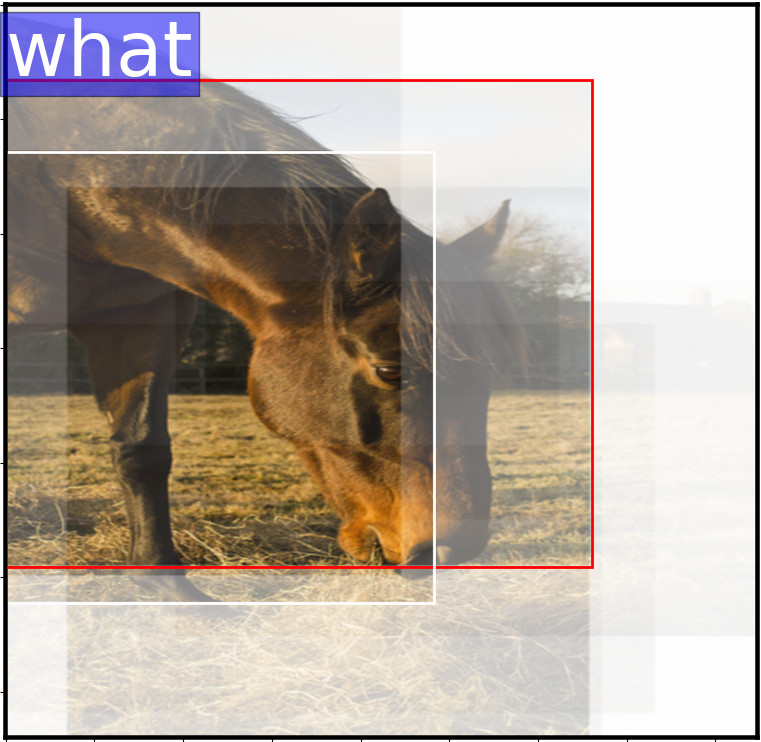}
            \caption{what}
            \label{fig:SRl}
    \end{subfigure}%
    \begin{subfigure} [b]{0.160\textwidth}        
            \includegraphics[width=1.0in,height=0.88in]{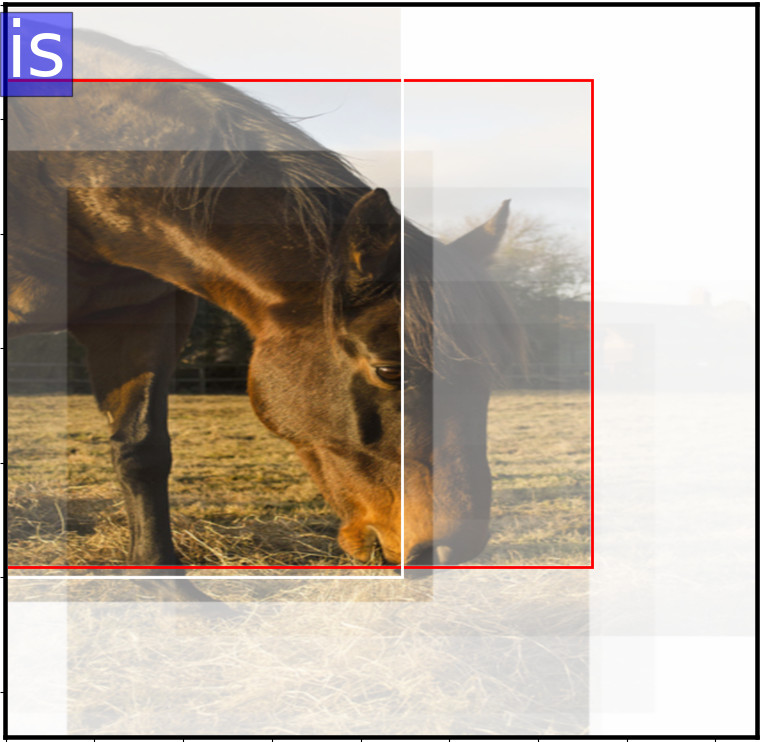}
            \caption{is}
            \label{fig:SRl}
    \end{subfigure}%
    \begin{subfigure} [b]{0.160\textwidth}        
            \includegraphics[width=1.0in,height=0.88in]{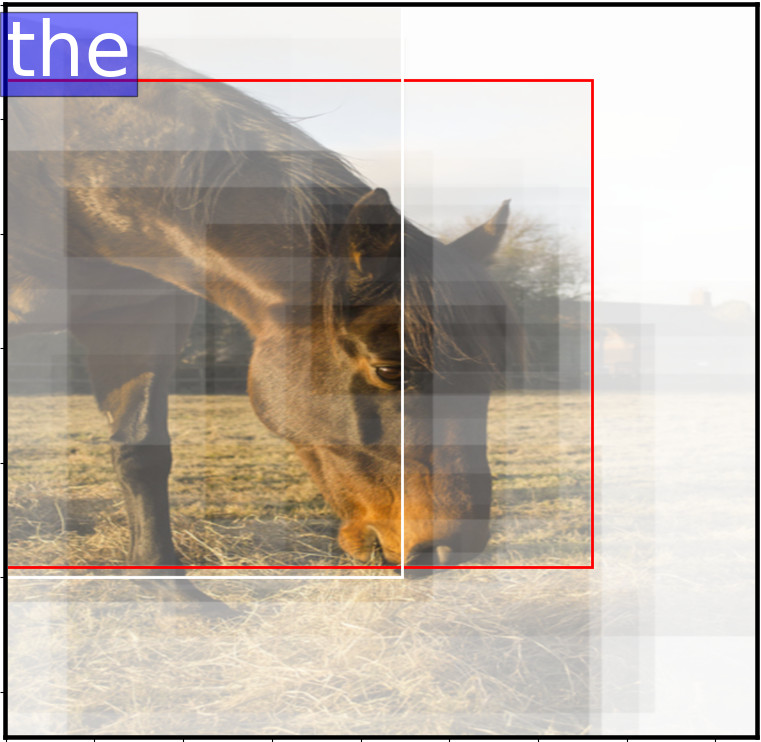}
            \caption{the}
            \label{fig:SRl}
    \end{subfigure}%
    \begin{subfigure} [b]{0.160\textwidth}        
            \includegraphics[width=1.0in,height=0.88in]{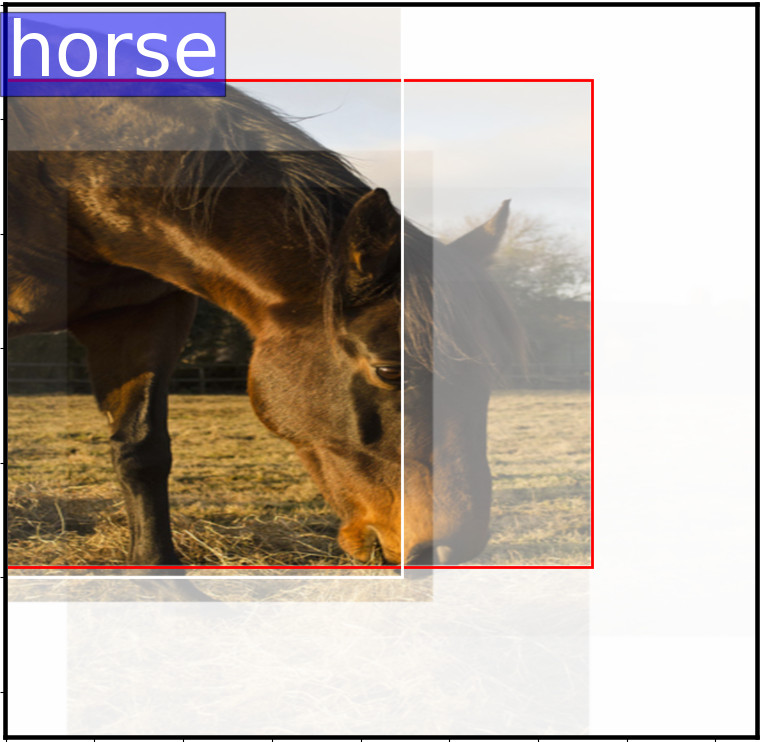}
            \caption{horse}
            \label{fig:SRl}
    \end{subfigure}%
    \begin{subfigure} [b]{0.160\textwidth}        
            \includegraphics[width=1.0in,height=0.88in]{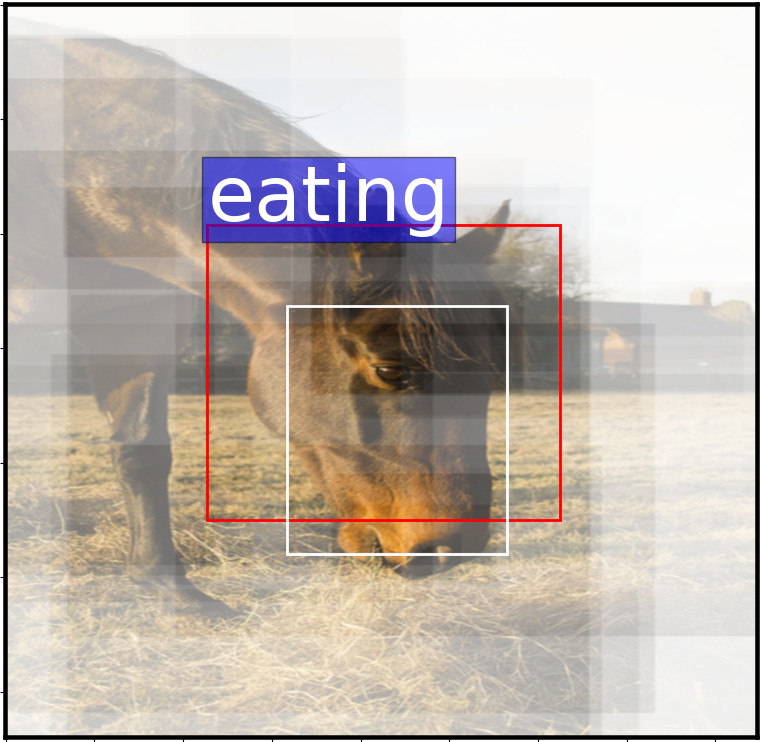}
            \caption{eating}
            \label{fig:SRl}
    \end{subfigure}%    
    \caption{Examples of question generation process by our model with corresponding dynamic attention visualization.
    The red and white bounding boxes localize regions with the largest and second largest attention weights respectively. 
    White regions have very low attention weights.
    }
    \label{qualitativeResults}
    \vskip -.2in
\end{figure*}
%%%%%%%%%%%%%%%%%%%%%%%%%%%%%%%%%%%%%%%%%%%%%%%%%%%%%%%%%%%%%%%%%%%
%------------------------------------------------------------------------- 

\subsection{Qualitative Analysis} 
To illustrate the function of our model's attention mechanism in generating corresponding questions 
for given image-answer pairs, 
we demonstrate a few examples of the question generation process of our proposed approach
with dynamic attention visualization 
in Figure \ref{qualitativeResults}.
The red box marks the salient region with the largest attention weight,
while the white box marks the salient region with the second largest attention weight, according to
Eq.(\ref{dattw}).

For the first two examples, we used the same image but with two different answers belonging to the same category (colour) 
to check the effectiveness of our proposed multi-level attention mechanism in capturing the fine-grained visual information based on a given answer. 
We can see that 
the proposed model generates very suitable corresponding questions for the given answers. 
Importantly, 
through our attention mechanism, the answer cue guide the model 
to attend to the exact corresponding objects (\eg the green sign in the first example and red hydrant in the second). 

In the third example, we used a more complex answer, specifically "turban". 
We notice that our model can semantically understand that the turban represents an object that people wear on their heads. 
Consequently, we can see our model tends to ignore most parts of the image and only focuses 
on those objects corresponding  to the currently predicted word (\eg, {\em man, wearing, head}).
Similarly in the last example, the model focuses on the horse object 
while generating the word ``{\em horse}" and 
on the horse' head and mouth while generating the word ``{\em eating}.  
These examples demonstrate that our approach has the capacity to correlate a given answer 
with specific corresponding regions in a given image. 
Meanwhile, it also has the capacity to dynamically change its attention according to the previously generated word.
These results again validated the effectiveness of the proposed model.

Another issue that is worth mentioning is that 
for a given image-answer pair, intuitively there could be multiple reasonable questions. 
For example, in the first example in Figure \ref{qualitativeResults},
given the answer ``{\em green}", one could ask ``What color is the tree leaves".
However, although this question makes sense, it misses the focus of the image,
while our model generates the question focusing on the more salient region.

\begin{figure}[h]
\vskip .45in
\captionsetup[subfigure]{font=small}
\centering
      \begin{subfigure}{0.22\textwidth}
        \includegraphics[width=1.0in,height=0.85in]{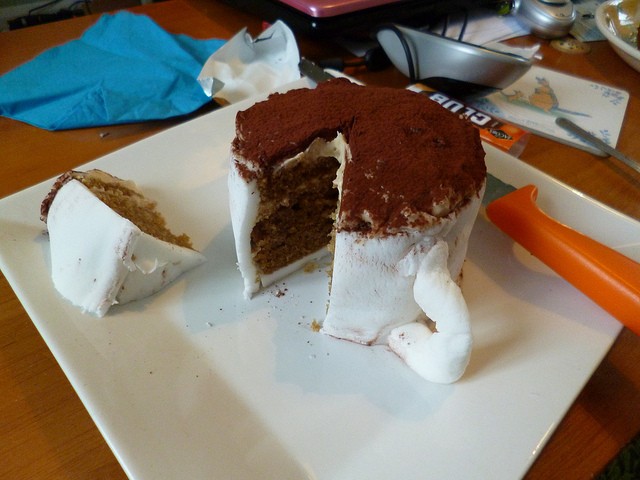}
          \label{div-yes}
      \end{subfigure}
      \begin{subfigure}{0.25\textwidth}
          \caption{Answer: yes\\\\\textbf{Is this a dessert?}\\Is this a cake?\\Is there a cake?
}
          \label{div-yes}
      \end{subfigure}
	 \\
      \begin{subfigure}{0.22\textwidth}
        \includegraphics[width=1.0in,height=0.85in]{chineese-diverse}
       
          \label{div-grass}
      \end{subfigure}
      \begin{subfigure}{0.25\textwidth}
          \caption{Answer: chinese\\\\\textbf{What type of cuisine is this?}\\What kind of cuisine is this?\\What type of food is this?}
          \label{div-grass}
      \end{subfigure}
      \caption{Examples of the diverse questions generated by our proposed model.
	 Questions in bold refer to the top questions selected by our model.}
 	\label{diversity}
\vskip -.4in
\end{figure}

Nevertheless, given an image and a very short answer phrase, 
a human can ask multiple questions with similar qualities.
To examine the capacity of the proposed model in generating diverse questions
for the same given image and answer pair, 
we used a beam size of 4 to perform inference and present the top 3 questions
with the highest probability scores. 
In Figure \ref{diversity}, we presented two examples.
We can see that the short answer phrase admits 
enormous possibilities for generating corresponding questions.
The top questions generated by our model all seem to be very reasonable,
while focusing on the salient objects.

%-------------------------------------------------------------------------
\section{Conclusion}\label{Conclusion}

In this paper, we proposed a novel multi-level attention model to tackle the new task of iVQA,
which has broader capacity for interacted visual and textual understanding than VQA.
Different from existing iVQA approaches, we enhance the visual representation extracted at the level of objects with their corresponding semantic information to reduce the semantic gap between visual and textual modalities. 
In addition, we deploy a dual guiding attention to help the model focus 
on the 
parts corresponding to a given answer 
and use a dynamic attention to attend to the most relevant image visual features in each recurrent question generation context. 
We conducted experiments on the VQA V1 dataset to compare the proposed model with existing iVQA methods.
The results show the effectiveness of the proposed approach by demonstrating state-of-the-art performance.

%-------------------------------------------------------------------------

{\small
\bibliographystyle{ieee}
\bibliography{paperbib}
}

\end{document}